\documentclass[10pt,twocolumn,letterpaper]{article}

\usepackage{cvpr}
\usepackage{times}
\usepackage{epsfig}
\usepackage{graphicx}
\usepackage{amsmath}
\usepackage{amssymb}
\usepackage{array,multirow}
\usepackage{booktabs}                  

\usepackage[draft,breaklinks=true,bookmarks=false]{hyperref}

\cvprfinalcopy 

\def\cvprPaperID{****} 

\ifcvprfinal\fi
\begin{document}

\title{Egocentric Human Segmentation for Mixed Reality }

\author{Andrija Gajic\\
Universidad Autonoma de Madrid\\
\and
Ester Gonzalez-Sosa \\
Nokia Bell-Labs\\
\and
Diego Gonzalez-Morin\\
Nokia Bell-Labs\\
\and
Marcos Escudero-Vi\~{n}olo\\
Universidad Autonoma de Madrid\\
\and
Alvaro Villegas\\
Nokia Bell-Labs\\
}

\maketitle

\begin{abstract}
  The objective of this work is to segment human body parts from egocentric video using semantic segmentation networks.~Our contribution is two-fold:~$i)$ we create a semi-synthetic dataset composed of more than $15,000$ realistic images and associated pixel-wise labels of egocentric human body parts, such as arms or legs including different  demographic factors; $ii)$ building upon the ThunderNet architecture, we implement a  deep learning semantic segmentation algorithm that is able to perform beyond real-time requirements ($16$ ms for $720\times720$ images). It is believed that this method will enhance sense of presence of Virtual Environments and will constitute a more realistic solution to the standard virtual avatars.
\end{abstract}

\section{Introduction}

In the past recent years, the release of video see-through cameras such as Microsoft Real Sense or ZED mini attached to headsets or built-in such as the ones in HTC Vive Pro have led Mixed Reality (MR) researchers to see the opportunities of egocentric vision. Augmented Virtuality (AV) is a MR subcategory that aims to augment a virtual environment (VE) with the reality surrounding him, previously captured with an egocentric camera. Among its main benefits, AV allows physical interaction with real objects and increases the awareness with the real world while being immersed in a VE. 


Users' sense of presence (SoP) within a VE is a very important construct that affects their propensity to experience the VE as if it real. One effective way to increase this SoP is by providing a user representation, which helps to shift from being a merely observer to really experiencing the VE \cite{slater1993influence}. Self-avatars are the mainstream solution followed in the Virtual Reality (VR) community, which are virtual models with a human-like appearance, and are mostly focused on hand representations. Apart from increasing the SoP, self-avatars also increase the sense of embodiment (SoE) \cite{argelaguet2016role}\footnote{As defined by \textit{Kilteni et al.} sense of embodiment refers to the feeling of being inside, controlling and having a virtual body.} and distance estimation \cite{mcmanus2011influence}.~Among their limitations, they have some problems related to misalignment between virtual and real body \cite{ogawa2020effect}. 

One promising application of egocentric vision for MR is the use of video self-avatars, that is, rather than using virtual hand models, adopting the user's real ones by segmenting the egocentric vision.~The VR community has explored this idea for some time. For instance color-based approaches \cite{fiore2012towards,bruder2009enhancing,immersirve_gastronomic2019,gunther2015aughanded} can be deployed in real time but tend to work well just for controlled conditions. However, they failed at dealing with different skin colors or with long-sleeve clothes \cite{fiore2012towards}.~Alternatively, depth solutions based on the incorporation of real objects within a distance from the camera have been proposed \cite{rauter2019augmenting}. Despite effective for some situations, they still lack enough precision to provide a generic, realistic and real time immersive experience. In our previous work \cite{gonzalez2020enhanced}, we explore deep semantic segmentation networks to perform arm segmentation using the EgoArm semi-synthetic dataset. Results proved their increased robustness against uncontrolled scenarios with respect to color-based or depth-based approaches, but the particular architecture explored was not light enough for real-time performance. Futhermore, the immersive experience and counterpart SoP and SoE can be further enhanced by providing not only video self-avatars of hands but of the entire human body. Therefore, this paper investigate how to integrate egocentric user's body from the egocentric capture in real time.

The rest of this article is structured as follows.~Section~\ref{related_works} discusses relevant related works.~In Section~\ref{egobody_dataset} we present the Egocentric Human Segmentation dataset created for this task.~Section~\ref{semantic_segmentation} describes the deep learning architecture designed to target real time egocentric segmentation.~Finally Section~\ref{results} reports some preliminary results and concludes the paper with some open problems and future works.

\section{Related Works}
\label{related_works}

In the literature there are several works that introduces users' whole body into the VE. For instance, Bruder \textit{et al.}~proposed skin segmentation algorithm to incorporate users' hands handling different skin colors \cite{bruder2009enhancing}. Conversely, a floor subtraction approach was developed to incorporate users' legs and feet  in the VE. Making the assumption that the floor appearance was uniform, the body was retained by simply filtering out all pixels not belonging to the floor \cite{bruder2009enhancing}. 

Then, Chen \textit{et al.}, in the context of 360$^{\circ}$ video cinematic experiences,~explored~depth~keying techniques to incorporate all objects below a predefined distance threshold \cite{chen2017effect}.~This distance threshold could be changed dynamically to control how much of the real world was shown in the VE. The user could also control the transitions between VE and real world through head shaking and hand gestures. Some of the limitations that the authors pointed out were related to the limited field of view of the depth sensor.

Pigny et Dominjon \cite{pigny2020using} were the first to propose a deep algorithm to segment egocentric bodies. Their architecture was based on U-NET and trained using a hybrid dataset composed of images from the COCO dataset belonging to persons and a $1500$-image custom dataset created following the automatic labelling procedure reported in \cite{gonzalez2020enhanced}. They reported $16$ ms of inference time for $256\times256$ images, and also observed problems of false positives that downgrade the AV experience.~In this work we plan to extend our previous work \cite{gonzalez2020enhanced} and explore further this segmentation problem targeting to perform real-time segmentation at higher resolutions, which are needed for creating and achieving a realistic immersive experience.
\section{Egocentric Body Dataset}
\label{egobody_dataset}

\begin{figure}[t]
  \centering
  \includegraphics[width=\linewidth]{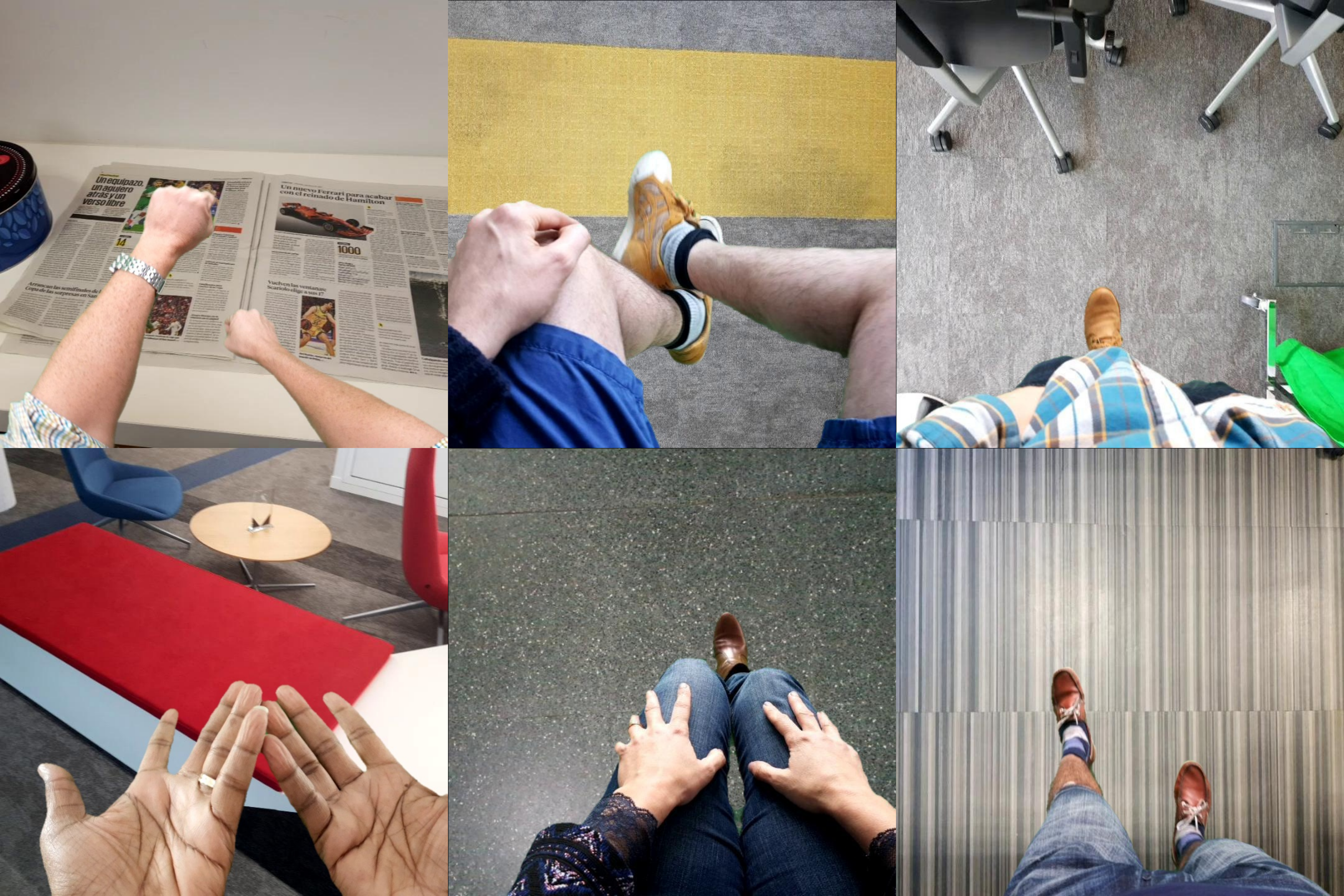}
  \caption{Samples from the Egocentric Human Segmentation Dataset: From Left to Rigth: stand up position looking to the front or slightly leaned, stand up position looking to the floor and sit down position looking to the floor. }
	\label{ego_human}
\end{figure}	

Due to the lack of egocentric human datasets with pixel-wise labelling, we decided to create our own, following the same procedure as reported in \cite{gonzalez2020enhanced}\footnote{As far as we know, the dataset created in \cite{pigny2020using} is not publicly available.}.~This procedure aimed to create semi-synthetic images by firstly capturing human body parts in front of chroma-key backdrop and then merging them with realistic backgrounds.~This smart method prevented us from the extremely time consuming and error-prone task of pixel-wise labelling. For this new task, we have extended our previously published EgoArm dataset with images from the egocentric lower-body parts, as it follows:

\begin{figure*}[th]
  \centering
  \includegraphics[width=0.7\linewidth]{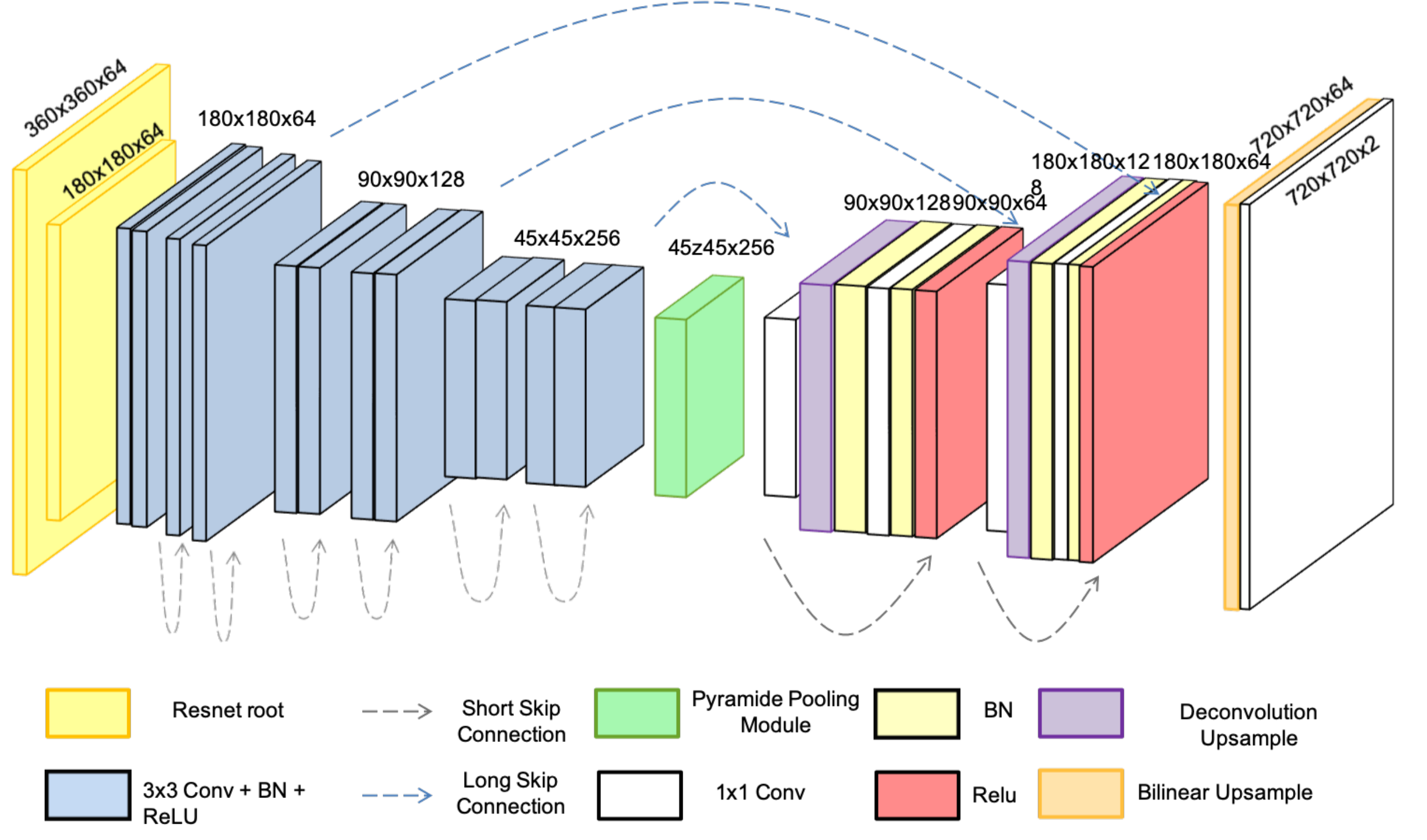}
  \caption{Deep Learning Arquitecture proposed for Egocentric Human Segmentation.}
	\label{deep_architecture}
\end{figure*}	

\begin{itemize}

\item \textbf{Egocentric body capture}: we asked a total of $13$ users to walk freely through the chroma-key backdrop while being recorded. The second round was performed with the users sitting down in a chair also covered by the chroma-key. The recording was done using an Android app installed in a Samsung S8 smartphone placed in the Samsung Gear Framework headset, taking $720\times720$ images at $30fps$. Users repeated the experiments with different outfits including short and long sleeves. Among the users, there is also variety in terms of gender, and skin color. Then, images were sampled from videos so that all users were equally represented.
\item \textbf{Egocentric background capture}: at the second stage, videos of realistic backgrounds were acquired using the same app in three different positions: at stand up position looking to the front; at stand up position looking to the floor, and at a sit-down position looking to the floor. A total of $73$, $27$ and $18$ different background videos were acquired for the different positions, encompassing different indoors scenarios including offices, houses, restaurants, and halls. $2$ frames were sampled per each video. Frames pertaining to the videos looking to the floor were augmented using rotation of $45^{\circ}$, $90^{\circ}$ and $180^{\circ}$. 

\item \textbf{HSV filtering and Alpha-channel based combination}: First, binary foreground masks of the recorded chroma-key videos are estimated using the process described in \cite{gonzalez2020enhanced}. Then, binarized foreground masks were smoothed by applying the Shared Sampling alpha matting algorithm \cite{gastal2010shared}.~Finally, alpha masks were used to realistically merge the segmented body parts with a randomly chosen background counterpart. Examples of resulting images can be seen in Fig. \ref{ego_human}.
\end{itemize}


A total of $6733$ lower body images were created. Before joining both datasets, EgoArm was down sampling from $17233$ images to $8668$ images, so both datasets were equally represented. Futhermore, the MIT Scene Parsing backgrounds used originally in the EgoArm dataset were replaced with the more realistic ones acquired in this new setup. As a result, we have a total of $15,401$ images conforming the Ego Human Segmentation dataset\footnote{It will be publicly available for research purposes for the camera ready version.}.

\section{Egocentric Deep Segmentation}
\label{semantic_segmentation}

Fig. \ref{deep_architecture} depicts the general architecture designed to segment among two classes: background and human egocentric body parts. It is inspired by the ThunderNet architecture \cite{xiang2019thundernet}. This architecture is mainly based on three parts: $1)$ an encoding subnetwork; $2)$ a pyramid pooling module (PPM); and $3)$ a decoding subnetwork. The encoding subnetwork is based on the first three Resnet-18 blocks \cite{xiang2019thundernet}. The output of the encoding block is followed by a PPM. Unlike the original network \cite{xiang2019thundernet} and due to the larger size of training images, we decided to use larger sampling pooling factors: $6,12,18,24$. The decoding subnetwork is similar to the one proposed in the original architecture, made up of two deconvolutional blocks. Besides, apart from the skip connections included within the encoding and decoding blocks, we include three more long skip connections between encoding and decoding subtnetworks for refining object boundaries. 

\subsection{Training and experimental protocol}

This new Thundernet architecture has been developed and trained using Keras framework. The weights from the three Resnet-18 blocks inside encoder are inherited from a model pre-trained on ImageNet dataset. Afterwards, the whole architecture is fine-tuned in an end-to-end approach. 
Among the entire Ego Human dataset, a total of $2743$ images were selected as the validation subset, assuring that users from both sets were disjoint. The remainder of images, a total of $12658$ were used as training. Different strategies were applied. First, the total set of backgrounds without foreground were included as part of the training set. Chromatic and cropping augmentation techniques were also applied to the training images. The loss function used was the weighted cross entropy, whose weights were estimated according to the whole frequency of foreground and background pixels in the training set ($0.56$ and $3.27$ for the background and human class, respectively). After many extensive experiments, the hyper-parameters found for the best performance were obtained using an Adam optimizer, a batch size of $8$ (due to the high size of the training images), $25$ epochs, learning rate of $1e-5$, and weight decay of $2e-4$. 



\section{Results and Conclusions}
\label{results}

\begin{table*}[]
\centering
\scriptsize
\begin{tabular}{|ccccccccccc|}
\hline
                                             & GTEA & EDSH2 & EDSHK & Ego Hands & EgoGesture & THU-READ & TEgO vanilla & \begin{tabular}[c]{@{}l@{}}TEgO vanilla\\ illu\end{tabular} & TEgO wild & \begin{tabular}[c]{@{}l@{}}TEgO wild\\ ilu\end{tabular} \\
                                             \hline
\cite{gonzalez2020enhanced} & $0.60$ & $0.74$  & $0.56$  & $0.33$      & $0.67$       & $0.48$     & $0.46$         & $0.48$                                                        & $0.37$      & $0.54$  
                                              \\
                                              \hline    
Ours                              & $0.54$ & $0.50$  & $0.40$  & $0.28$      & $0.61$       & $0.36$     & $0.27$         & $0.39$                                                        & $0.39$      & $0.53$                                                    \\
    \hline 
    \end{tabular}
      \caption{Empirical result comparison with the ones reported in our previous work \cite{gonzalez2020enhanced}, in terms of Intersection over Union in the range $0-1$}
      \label{fig:IoU}
\end{table*}

\begin{figure*}[t]
  \centering
  \includegraphics[width=0.8\linewidth]{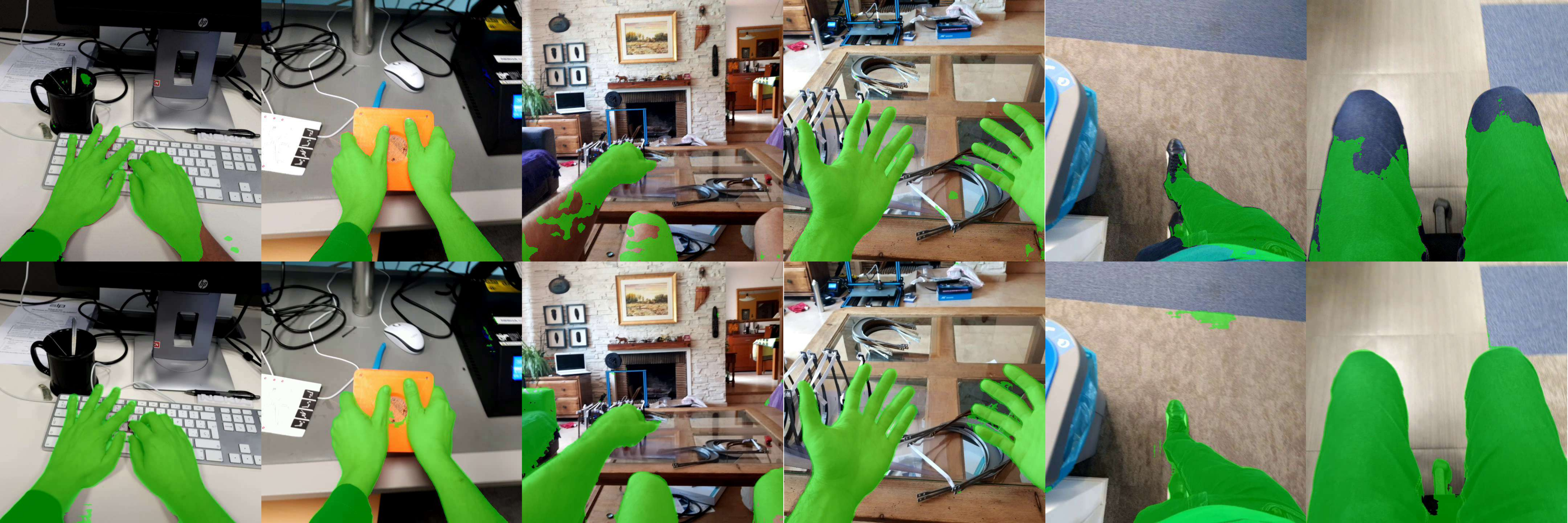}
  \caption{Qualitative results from real egocentric frames. Above, results achieved with our previos method \cite{gonzalez2020enhanced}, below, results achieved with the new and shallower Thundernet arquitecture.}
	\label{fig:results}
\end{figure*}	

Table \ref{fig:IoU} reports the Intersection over Union from different egocentric arm/hand datasets. As can be seen, in most cases, the new network is achieving comparable results using a shallower arquitecture (notice that due to some discrepancy between groundtruth associated to these datasets, some results are underestimated, see \cite{gonzalez2020enhanced} for clarification). Another benefit of this new network, is their ability to segment whole egocentric bodies. Due to the lack of datasets with labelled human body, Fig. \ref{egobody_dataset} presents some qualitative results from real egocentric $720\times720$ frames. The most interesting benefit is that, unlike previous approaches \cite{pigny2020using}, our method is able to perform real time segmentation on high resolution images (inference time per $720\times720$ images takes only $16ms$ using a PC Intel Xeon ES-2620 V4 @ $2.1$Ghz with $32$ GB powered with $2$ GPU GTX-1080 Ti with $12GB$ RAM. 

We experience that without the smoothed blending between human body parts and realistic backgrounds, the network tends to focus the learning mainly on the edges and could not provide good segmentation. Second, the use of chromatic  and cropping augmentation helps the network to be less dependent of illumination and spatial position, respectively. 
The weighted cross entropy loss helps to focus the learning more on the human body parts, rather than in the backgrounds, whose unlimited variability can not be completely retained in the network. Despite the good segmentation accuracy and real time performance, we observed that there are still some false positives in the background. For future work, we plan to tackle more in depth this problem by studying more sophisticated loss functions, or approaching the training in two stages, and train as a first step the arquitecture on a bigger dataset with more background classes.


%
%

{\small
\bibliographystyle{ieee_fullname}
\bibliography{egbib}
}

\end{document}